\documentclass[10pt,twocolumn,letterpaper]{article}

\usepackage{wacv}
\usepackage{times}
\usepackage{epsfig}
\usepackage{graphicx}
\usepackage{amsmath}
\usepackage{amssymb}
\usepackage{booktabs}
\usepackage{comment}
\usepackage{color}
\usepackage{graphicx}
\usepackage{amsmath}
\usepackage{amssymb}
\usepackage{booktabs}
\usepackage{multicol}
\usepackage{multirow}
\usepackage{subfigure}
\usepackage{colortbl}
\usepackage{enumitem}
\usepackage[normalem]{ulem}
\usepackage[T1]{fontenc}
\usepackage{dsfont}
\usepackage{ragged2e}
\usepackage{cite}
\usepackage{amsmath}
\usepackage{floatrow}
\usepackage{blindtext}
\usepackage{balance}

\makeatletter
\@namedef{ver@everyshi.sty}{}
\makeatother
\usepackage{tikz}

\newcommand{\formattedparagraph}[1]{\noindent \textbf{#1}}
\newcommand{\etals}{\textit{et al.}}
\definecolor{RoyalBlue}{cmyk}{1, 0.50, 0, 0}

\wacvfinalcopy 

\ifwacvfinal
\usepackage[breaklinks=true,bookmarks=false,colorlinks=true, pagebackref=true]{hyperref}
\else
\usepackage[pagebackref=true,breaklinks=true,colorlinks,bookmarks=false]{hyperref}
\fi

\begin{document}

\title{Multi-View Photometric Stereo Revisited}

\author{\quad Berk Kaya$^{1}$\quad Suryansh Kumar$^{1}\thanks{Corresponding Author ({\tt\small k.sur46@gmail.com})}$ \quad Carlos Oliveira$^1$ \quad Vittorio Ferrari$^{2}$\quad Luc Van Gool$^{1, 3}$\\
ETH Z\"urich${^1}$, Google Research$^2$, KU Leuven$^3$
}

\pdfoutput=1


\maketitle
\thispagestyle{empty}

\begin{abstract}
Multi-view photometric stereo (MVPS) is a preferred method for detailed and precise 3D acquisition of an object from images. Although popular methods for MVPS can provide outstanding results, they are often complex to execute and limited to isotropic material objects. To address such limitations, we present a simple, practical approach to MVPS, which works well for isotropic as well as other object material types such as anisotropic and glossy. The proposed approach in this paper exploits the benefit of uncertainty modeling in a deep neural network for a reliable fusion of photometric stereo (PS) and multi-view stereo (MVS) network predictions. Yet, contrary to the recently proposed state-of-the-art, we introduce neural volume rendering methodology for a trustworthy fusion of MVS and PS measurements. The advantage of introducing neural volume rendering is that it helps in the reliable modeling of objects with diverse material types, where existing MVS methods, PS methods, or both may fail. Furthermore, it allows us to work on neural 3D shape representation, which has recently shown outstanding results for many geometric processing tasks. Our suggested new loss function aims to fits the zero level set of the implicit neural function using the most certain MVS and PS network predictions coupled with weighted neural volume rendering cost. The proposed approach shows state-of-the-art results when tested extensively on several benchmark datasets.
\end{abstract}

%

\section{Introduction}
Multi-view photometric stereo (MVPS) aims at recovering accurate and complete 3D reconstruction of an object using multi-view stereo (MVS) and photometric stereo (PS) images \cite{hernandez2008multiview}. While PS is exemplary in recovering an object's high-frequency surface details, MVS helps in retaining the global consistency of the object's 3D shape and assists in correcting overall low-frequency distortion due to PS \cite{nehab2005efficiently,furukawa2015multi, kaya2022uncertainty}. Hence, MVPS inherits the complementary output response of PS and MVS methods. Contrary to the active range scanning methods \cite{nehab2005efficiently,chatterjee2013efficient, sandstrom2022learning}, it provides an efficient, low-cost, and effective alternative for trustworthy 3D data acquisition. And therefore, it is widely preferred in architectural restoration \cite{nehab2005efficiently}, machine vision industry \cite{hernandez2008multiview, kaya2022uncertainty, sarno2022neural}, etc.
%

State-of-the-art {geometric methods} to solve MVPS indeed provide accurate results but are composed of multiple optimizations and filtering steps applied in sequel \cite{hernandez2008multiview,li2020multi,park2016robust}. Further, these steps are intricate and require the manual intervention of an expert for precise execution, thereby limiting its automation \cite{li2020multi,park2016robust}. Moreover, these approaches cannot meet modern industrial requirements of scalability and low-memory footprint for efficient storage of recovered 3D models. Lately, neural network-based {learning methods} to solve MVPS have shown few critical advantages over geometric methods \cite{kaya2022uncertainty, kaya2020uncalibrated}. These methods are simpler, effective, and can provide a high-quality 3D model with a lower memory footprint. Yet, they depend on specific assumptions about the material type, which limits their application to anisotropic and glossy material objects.

\begin{figure*}[t]
\centering
\subfigure[\label{fig:acquisition_setup} MVPS Setup ]{\includegraphics[width=0.35\linewidth]{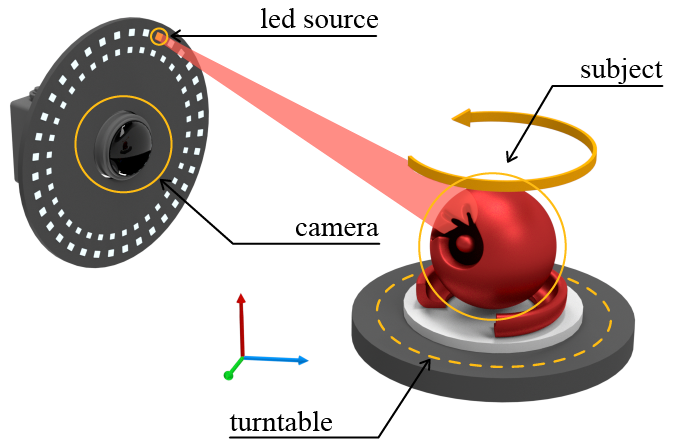}} ~~~~~~~~
\subfigure[\label{fig:anisotropic_brdf} Benefit of our approach]{\includegraphics[width=0.35\linewidth]{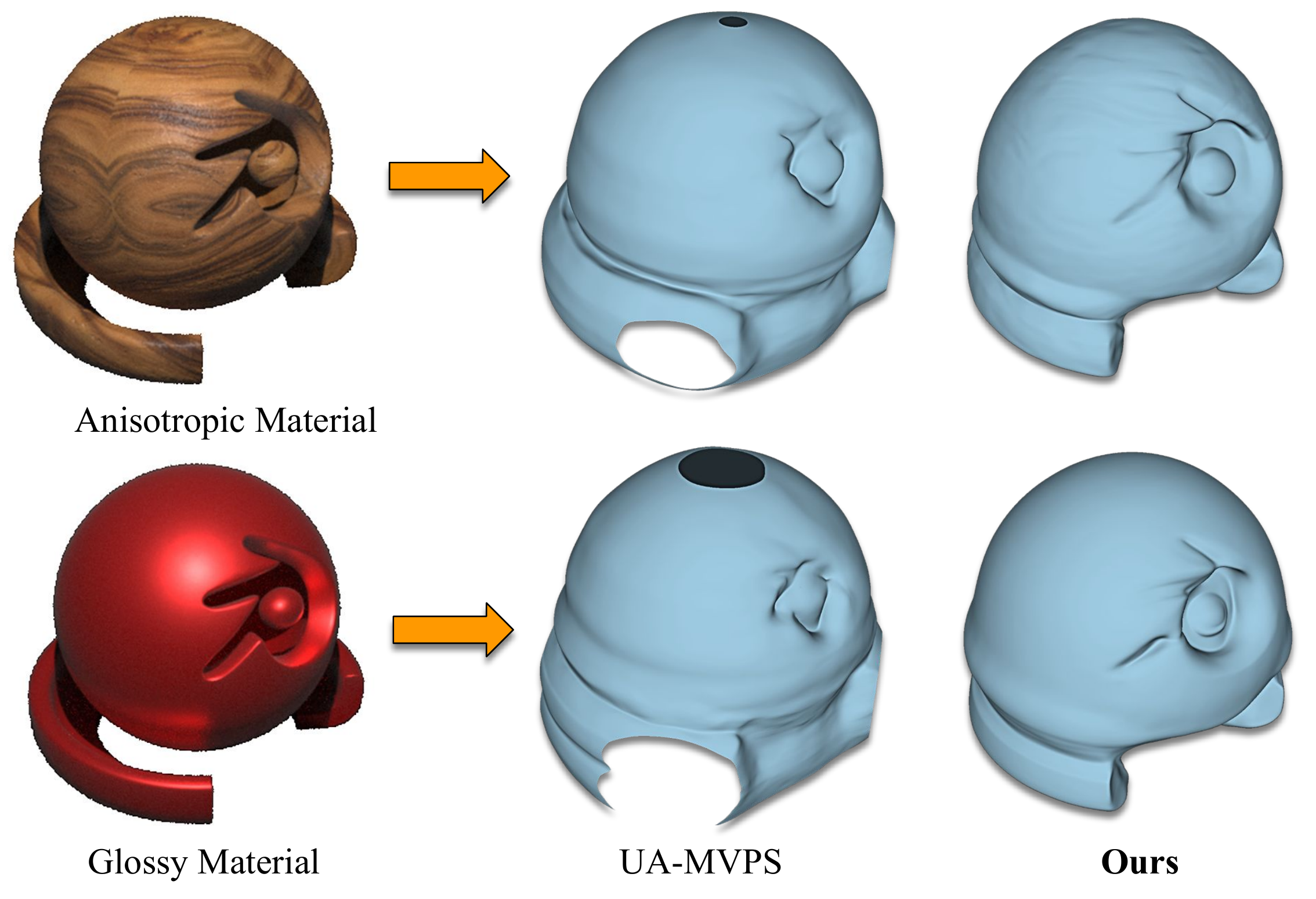}}
\caption{\small (a) The classical MVPS setup as outlined in Hern${\acute{\text{a}}}$ndez \etals  \cite{hernandez2008multiview} work. (b) The advantage of our method over current state-of-the-art deep-MVPS method \ie, UA-MVPS \cite{kaya2022uncertainty}. It can be observed that our method is able to correctly recover the fine object's details for anisotropic and glossy material object. The 3D model used for the above illustration is taken from \cite{mildenhall2020nerf} dataset.}
\label{fig:mvps_figure}
\end{figure*}

In this paper, we present a general yet simple and effective approach to the MVPS problem. Inspired by the recent MVPS method \cite{kaya2022uncertainty}, we introduce uncertainty modeling in multi-view stereo and photometric stereo neural networks for reliable inference of the 3D position and surface normals, respectively. Although uncertainty estimation helps us filter wrong predictions, it can lead to incomplete recovery of an object's 3D shape. To this end, Kaya et al. \cite{kaya2022uncertainty} recently proposed Eikonal regularization to recover the missing details due to filtering. On the contrary, we introduce neural volume rendering of the implicit 3D shape representation. It has couple of key advantage over \cite{kaya2022uncertainty} pipeline: \textbf{\textit{(i)}} It helps extending the application of MVPS to a wider class of object with different material type (see Fig.\ref{fig:anisotropic_brdf}). \textbf{\textit{(ii)}} It further enhances the performance and use of implicit neural shape representation in MVPS leading to state-of-the-art results on benchmark datasets.





Meanwhile, recent multi-view stereo approaches have shown that neural volume rendering using the implicit neural 3D shape representation can effectively model a diverse set of objects via multi-view image rendering techniques \cite{mildenhall2020nerf,yariv2021volume,yariv2020multiview,lee2022uncertainty, jain2022robustifying}. Therefore, introducing it to MVPS can assist in handling challenging objects' material types. Intuitively, rendering-based geometry modeling can succeed where both the MVS and PS methods fail to estimate the surface geometry \cite{li2020multi,nehab2005efficiently,park2016robust}. Further, contrary to the standard practice in MVPS of performing optimization or filtering on explicit geometric primitives \cite{nehab2005efficiently,li2020multi,park2016robust}, \ie, mesh, neural volume rendering relies on neural implicit shape representation, which is memory efficient and is scalable \cite{yariv2021volume}. In summary, our paper makes the following contributions:

\begin{itemize}[leftmargin=*, topsep=0pt, noitemsep]
    \item  We present a simple, efficient,  scalable, and effective MVPS method for the detailed and complete recovery of the object's 3D shape.
    \item Our proposed uncertainty-aware neural volume rendering uses confident priors from deep-MVS and deep-PS networks and encapsulates them with an implicit geometric regularizer to solve MVPS demonstrating state-of-the-art reconstruction results on the benchmark dataset \cite{li2020multi}.
    \item Contrary to the current state-of-the-art methods, our method applies to a broader class of object material types, including anisotropic and glossy materials. Hence, widen the use of MVPS for 3D data acquisition.
\end{itemize}


\section{Related Work}\label{sec.related_works}
\formattedparagraph{{Classical MVPS.}}
Early MVPS methods assume a particular analytic BRDF model, which may not be apt for real-world objects whose reflectance differs from the assumed BRDF model \cite{hernandez2008multiview, goldman2009shape, lim2005passive}. Later, Park \etals \cite{park2013multiview, park2016robust} proposed a piece-wise planar mesh parameterization approach for recovering an object's fine surface details via displacement texture maps. Nevertheless, their work was not aimed at modeling surface reflectance properties. Other methods such as \cite{ren2011pocket, dong2010manifold} model the BRDF, yet restricted to near-flat surface modeling assuming the surface normal is known.

Other classical MVPS methods that have been proposed in the last couple of years do provide decent results \cite{li2020multi, zhou2013multi}; yet, their introduced pipeline composes of several complex optimization algorithms such as iso-depth contour estimation, contour tracing, structure-from-motion, multi-view depth propagation, point sorting, mesh optimization using  \cite{nehab2005efficiently}, and ACLS algorithm \cite{lawrence2006inverse}. Moreover, some of these steps require an expert's intervention for parameter fine-tuning; hence challenging to re-implement, automate and execute. Additionally, the method's reflectance modeling is built on Alldrin \etals \cite{alldrin2008photometric} and Tan \etals \cite{tan2007isotropy} work, and therefore, its application is limited to isotropic material objects.

\smallskip
\formattedparagraph{{Deep MVPS.}}
In recent years, deep learning-based approaches to MVPS have been proposed as alternatives to classical methods. Not long ago, Kaya \etals \cite{kaya2021neural} introduced a neural radiance fields-based MVPS approach (NR-MVPS). The proposed pipeline predicts the object's surface normals using a deep-PS network and blends them in a multi-view volume rendering formulation to solve MVPS. Regardless of its simplicity, it fails to provide a high-quality 3D reconstruction of the object. Further, \cite{kaya2020uncalibrated} proposed neural inverse rendering idea  to recover an object's shape and material properties. Among all the deep MVPS methods, the recently introduced uncertainty-based MVPS approach \cite{kaya2022uncertainty} (UA-MVPS) provides better 3D reconstruction results. However, it fails on anisotropic and glossy objects (see Fig.\ref{fig:anisotropic_brdf}). On the contrary, this paper proposes a method that can successfully make MVPS 3D acquisition setup work for isotropic, anisotropic, and glossy objects with magnificent results.



\section{Preliminaries}

\formattedparagraph{MVPS Setup.} Hern${\acute{\text{a}}}$ndez \etals \cite{hernandez2008multiview} proposed the introductory MVPS acquisition setup\footnote{Refer Nehab et al. \cite{nehab2005efficiently} 2005 work, which uses active range scanning sensor to solve a similar problem.}. It is composed of a turntable arrangement, where light-varying images (PS images) of the object placed on the table are captured from a given viewpoint. Note that the camera and light sources' position remains fixed, and only the table rotates, providing a new viewpoint ($v$) of the object per rotation. For every table rotation, PS images for each light source are captured and stored (see Fig.\ref{fig:acquisition_setup}).

\smallskip
\formattedparagraph{Notation and Definition.} Denoting $L$ as the total number of point light sources and $V$ as the total number of viewpoints (corresponds to each table rotation), we define $\mathcal{X}_{ps}^{v} = \{X_{1}^v, X_{2}^v..., X_{L}^v\}$ as the set of photometric stereo images from each viewpoint $v \in [1, V]$, and $\mathcal{Y}_{mv} = \{Y^1, Y^2..., Y^{V}\}$ as the set of multi-view images constructed using $Y^{v} = \text{median} (\mathcal{X}_{ps}^v)$ as performed in \cite{li2020multi}.  The goal of an MVPS algorithm under calibrated setting is to recover the precise and complete geometry of the object. The motivation for using MVS and PS is due to the observation elaborated in \cite{nehab2005efficiently}. As alluded to above, despite PS can provide reliable high-frequency geometric details, it generally contributes to low-frequency surface distortion at coarse scale \cite{nehab2005efficiently}. We can correct such distortions using geometric constraints with object's MVS images.

Using the basic MVPS experimental setup, it is easy to recover two types of surface priors: \textit{(i)} 3D position per pixel ($\mathbf{p}_{i}\in\mathbb{R}^{3 \times 1}$) of the object using multi-view stereo images \textit{(ii)} surface normal for each surface point ($\mathbf{n}_{i}^{ps} \in \mathbb{R}^{3 \times 1}$) using light varying images \cite{woodham1980photometric, furukawa2015multi, hartley2003multiple, kutulakos2000theory}\footnote{Note that MVS reconstruction may not provide reliable per pixel 3D reconstruction. Hence, bad 3D estimates are filtered which leads to sparse set of object 3D points.}. Hence, by design, the problem boils down to effective use MVS and PS surface priors, light varying images, light and camera calibration data for high-quality dense 3D surface recovery. To have the 3D position prior, most methods resort to structure from motion method or its variation \cite{park2013multiview, li2020multi}. For surface normal prior, one of the popular image formation model is:
\begin{equation}
    \begin{aligned}\label{eq:generalPS}
       {X}_{j}^{v}(\mathbf{p}_i) = {e}_{j}\cdot\rho\big(\mathbf{n}_i(\mathbf{p}_i), \mathbf{l}_{j}, \mathbf{v}\big) \cdot \zeta_a\big(\mathbf{n}_i(\mathbf{p}_i), \mathbf{l}_{j}\big) \cdot \zeta_c(\mathbf{p}_i)
    \end{aligned}
\end{equation}
Here, the function $\rho()$ denotes the BRDF,  $\zeta_a (\mathbf{n}_i(\mathbf{p}_i), \mathbf{l}_{j}) = \max(\mathbf{n}_i(\mathbf{p}_i)^{T}\mathbf{l}_{j}, 0)$ accounts for the attached shadow, and $\zeta_c(\mathbf{p}_i) \in \{0, 1\}$ assigns $0$ or $1$ value to $\mathbf{p}_i$ depending on whether it lies in the cast shadow region or not. $\mathbf{l}_{j}$ is the light source direction and ${e}_{j} \in \mathbb{R}_+$ is the scalar for light intensity value due to $j^{th}$ light source. Although surface normal can be estimated with reasonable accuracy using Eq.\eqref{eq:generalPS} image formation model \cite{chen2019self}, modeling BRDF using it can be challenging. Therefore, we propose a neural network-based image rendering approach to overcome such a limitation. Experimental results show that using our approach help MVPS work for a broader class of object material. Next, we describe our approach to the MVPS problem in detail.



\section{Our Approach}\label{sec:method}
As mentioned in Sec.\ref{sec.related_works}, on the one hand, we have the state-of-the-art geometric method that is composed of several complex steps, hence not suitable for automation. Further, it cannot meet the modern demand of scalability, and thus, less convincing for the current challenge of handling a large set of object data. On the other hand, UA-MVPS \cite{kaya2022uncertainty} recent work on deep MVPS is simple and scalable but works well only for isotropic material objects. 

This paper proposes a simple, scalable, and effective approach that can handle a much broader range of objects. We first recover the 3D position and surface normal priors from MVS and PS images (MVPS setup) using uncertainty-aware deep multi-view stereo \cite{wang2021patchmatchnet} and deep photometric stereo networks \cite{ikehata2018cnn, kaya2022uncertainty}, respectively.  The uncertainty-aware network measures the suitability of the predicted surface measurements for its reliable fusion. However, the filtering of unreliable predictions based on the uncertainty measures leads to the loss of local surface geometry. Thus, we introduce a geometric regularization term in the overall loss function to recover the complete 3D geometry of the object. To that end, we represent the object's shape as level sets of a neural network and recover it by optimizing the parameters of a multi-layer perceptron (MLP). The MLP approximates a signed-distance-function (SDF) to a plausible surface based on the point cloud, surface normals, and an implicit geometric regularization term developed on the Eikonal partial differential equation \cite{crandall1983viscosity}.


%
%

The above pipeline is inspired by UA-MVPS \cite{kaya2022uncertainty}, which generally works well but cannot model anisotropic or glossy surfaces. Hence, not a general solution and is unsuitable for large applications. On a different note, we observed that representing the light fields and density of the object as a neural network in a multi-view volume rendering algorithm improves the 3D reconstruction of general objects. Further, as well-studied, volume rendering generalizes well to diverse objects with different material types. Such an observation lead us to introduce an uncertainty-aware volume rendering approach to the MVPS problem. As we will show, it not only helps achieve state-of-the-art results on isotropic material objects but also provide accurate 3D surface reconstruction on challenging subjects such as glossy texture-less surface objects. Next, we describe each component of our approach in detail, leading to the final loss.


\subsection{Uncertainty-Aware Deep-MVS Network}
Given a set of multi-view images $\mathcal{Y}_{mv}$,  $\{\mathbf{K}_v, \mathbf{R}_v, \mathbf{t}_v \}_{v=1}^{V}$ the set of camera intrinsics, rotations, and translations for each camera view, the goal is to recover the 3D position of the object corresponding to each pixel with a measure of its reconstruction quality. For that, we use PatchMatchNet \cite{wang2021patchmatchnet} architecture due to its state-of-the-art (SOTA) performance on large-scale images. Further, it provides dense depth maps with per-pixel confidence values. Such an inherent property allows the filtering of unreliable depth predictions without having to add an extra uncertainty estimation module into the network.



Built on the idea of classical PatchMatch \cite{barnes2009patchmatch} algorithm, it starts by generating random depth hypotheses. Then, the network repeatedly propagates and evaluates existing depth hypotheses at different image scales in a coarse-to-fine manner. Specifically, feature maps are extracted from each input image and the extracted features are used to generate new depth hypotheses. Subsequently, generated hypotheses are evaluated to compute the matching cost. For that, similarities between warped feature maps are calculated using group-wise correlation \cite{xu2020learning}. Finally, the depth $\mathbf{d}_i$ and the confidence $\mathcal{C}_i$ value at pixel $i$ are computed as follows:     
\begin{equation}\label{eq:depth regression}
    \mathbf{d}_i = \sum_{j=1}^{\mathcal{H}} d_i^j \cdot \textrm{softmax}(\mathbf{J}_i^j),   ~~~~  \mathcal{C}_i = \textrm{softmax}(\mathbf{J}_i^{j^*})
\end{equation}
Here, $d_i^j$ is the $j^{th}$ depth hypothesis at pixel $i$ and $\mathbf{J}_i^j$ is the computed matching cost of corresponding depth hypothesis. 
$\mathcal{H}$ is the total number of depth hypotheses, and $j^*$ is the most likely depth hypothesis at a pixel. After PatchMatchNet is applied at the finest image scale, 
we obtain the position estimate at pixel coordinates $\mathbf{o}_i$ by $\mathbf{p}_i = \mathbf{R}_{v}\big( \mathbf{d}_{i} \mathbf{K}_{v}^{-1} \mathbf{o}_i  \big) + \mathbf{t}_{v}$.  Further, we introduce per-pixel binary variable $\textcolor{blue!80!black}{c_i^{\text{mvs}}}$ to indicate highly confident estimates. We assign $\textcolor{blue!80!black}{c_i^{\text{mvs}}}=1$ when $\mathcal{C}_i > \tau_{\text{mvs}}$ and keep $\textcolor{blue!80!black}{c_i^{\text{mvs}}}=0$ for the rest \cite{kaya2022uncertainty}. For more details on deep-MVS network's train and test time specifics refer supplementary or \cite{wang2021patchmatchnet}.

\subsection{Uncertainty-Aware Deep-PS Network}
To predict surface normals per view from PS images $\mathcal{X}_{ps}^{v}$, 
and light source directions $\{\mathbf{l}_{j} \}_{j=1}^{L}$, we use the network architecture presented in \cite{ikehata2018cnn}. Instead of having a parametric BRDF model assumption, the network learns from training data to map an input \emph{observation map} to a surface normal. An observation map is a 2D matrix-based representation obtained by storing the intensity values at a pixel due to different light sources. Experiments suggest that observation map based representation facilitates accurate estimation of surface normals for general isotropic BRDFs \cite{zheng2019spline, yao2020gps}. For more details on the network architecture and observation map refer to Ikehata's work \cite{ikehata2018cnn} or supplementary material.

Despite the PS network architecture can predict the object's surface normals, it cannot measure uncertainty in the predicted value, which is one of the critical components of our approach. Following \cite{kaya2022uncertainty}, we adopt the Monte Carlo (MC) dropout approach \cite{ gal2015bayesian, gal2016dropout} and build up an uncertainty-aware deep-PS architecture. In a nutshell, we introduce a dropout layer with probability $p_{\text{mc}}$ after all convolution and fully connected layers. With this adjustment, the network can be treated as a Bayesian neural network, whose parameters approximate a Bernoulli distribution. Thus, we can train the network with an additional weight decay term scaled by $\lambda_{w}$ on network parameters: 
\vspace{-0.4cm}

\begin{equation}\label{eq:ps_network_loss}
  \mathcal{L}_{ps} = \frac{1}{{\mathbf{N}}_{\text{mc} }} \sum_{j=1}^{\mathbf{N}_{\text{mc}}} \lVert \mathbf{\tilde{n}}_j - \mathbf{n}_{\text{gt}} \rVert_2^2 + \lambda_{w} \sum_{k=1}^{K} \lVert \mathbf{W}_k \rVert_2^2
\end{equation}
In Eq.\eqref{eq:ps_network_loss} $\mathbf{\tilde{n}}_j$, $\mathbf{n}_{\text{gt}}$ denotes the network's predicted and ground-truth surface normal, respectively.
%
%
$\mathbf{N}_{\text{mc}}$ is the number of MC samples and $\mathbf{W}_k$ stands for the network weights at layer $k=1,...,K$. We train the network on CyclesPS dataset \cite{ikehata2018cnn} once, and used the same network for testing.

At test time, we keep dropout layers active to have a non-deterministic network and we run the network multiple times on the same input. This allows us to capture the fluctuation on the surface normal predictions. We average out all predictions at pixel $i$ to compute the output normal $\mathbf{n}_{i}^{ps} \in  \mathbb{R}^{3 \times 1}$ and the variance $\tilde{\sigma}_{i}^{2} \in \mathbb{R}^{3 \times 1}$. Since we are interested in highly confident predictions, we assign $\textcolor{green!70!black}{c_i^{\text{ps}}}=1$ if $\| \tilde{\sigma}_i^{2}\|_{1} < \tau_{\text{ps}}$ and keep $\textcolor{green!70!black}{c_i^{\text{ps}}}=0$ for the remaining pixels \cite{kaya2022uncertainty}. Here, $\textcolor{green!70!black}{c_{i}^{\text{ps}}}$ is a binary variable to indicate the selection of confident normal prediction.




\begin{figure*}[t]
    \centering
    \includegraphics[width=0.82\textwidth]{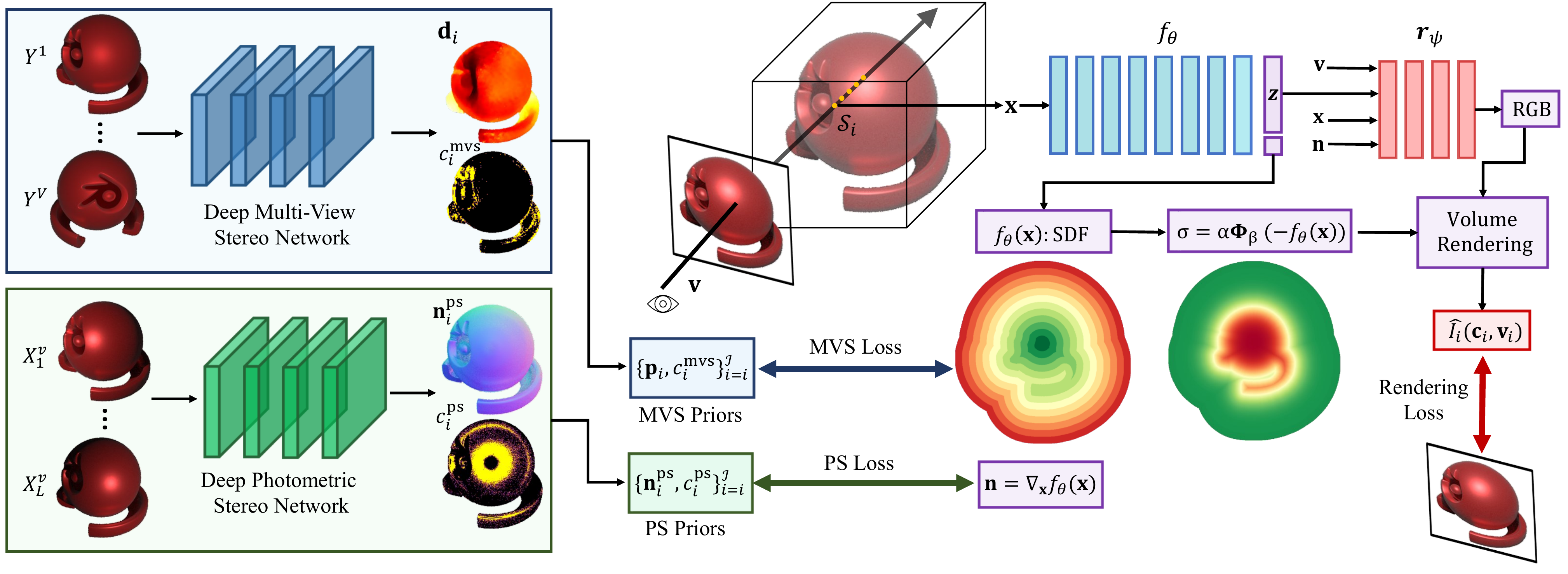}
    \caption{\textbf{Method overview (Left to Right)}: We obtain highly confident 3D position and surface normal predictions of the object via uncertainty-aware deep-MVS and deep-PS networks, respectively. Then, we learn the signed distance function representation of the object surface. Finally, our optimization uses the volume rendering technique to recover the missing details of the surface, providing high-quality 3D reconstructions of challenging material types.}\label{fig:pipeline}
\end{figure*}

\subsection{Shape Representation and Regularization}\label{sec:IGR}
Using deep-MVS and deep-PS networks ---as described above, we filter confident 3D positions and surface normals $\{ \mathbf{p}_{i}, \mathbf{n}_{i}^{ps} \}_{i=1}^{\mathcal{I}} \subset  \mathbb{R}^3$ prediction $ \forall i \in [1,\dots,\mathcal{I}]$. Our goal is to recover object's dense 3D reconstruction combining those reliable intermediate priors. To this end, we propose to learn the signed distance function (SDF) of the object surface defined by a implicit function $f_{\theta}(\mathbf{x}): \mathbb{R}^3 \rightarrow \mathbb{R}$ using the reliable prediction estimates. We model the function using an MLP parameterized by $\theta$, assuming its zero level set approximates the object surface. 
%
%
%

To find the optimal $\theta$, we consider the Eikonal equation ($||\nabla_{\mathbf{x}} f_{\theta}(\mathbf{x})|| = 1 $). It establishes a constraint on  $f_{\theta}(\mathbf{x})$ to represent a true SDF. Note that even if the boundary conditions imposed by the given surface estimates are satisfied (\ie, $f_{\theta}(\mathbf{p}_{i}) = 0$, $\nabla_{\mathbf{x}} f_{\theta}(\mathbf{p}_{i}) = \mathbf{n}_{i}^{ps}$), a unique solution to the zero level set surface may not exist. Nevertheless, describing an incomplete set of surface 3D estimates using Eikonal condition as a regularizer favors smooth and plausible surfaces  \cite{gropp2020implicit}. Hence, we consider the following regularization term in our optimization:

\begin{equation}\label{eq:loss_eikonal}
  \mathcal{L}_{\text{Eikonal}} = \lambda_e \mathbb{E}_{\mathbf{x}}  ( \lVert \nabla_{\mathbf{x}}f_{\theta}(\mathbf{x})\rVert -1 )^2
\end{equation}
where the expectation is computed w.r.t. a probability distribution $ \mathbf{x} \sim \mathcal{D}$.
Note that recent work \cite{kaya2022uncertainty} has considered the Eikonal regularization to interpolate the surface from MVS and PS network predictions. However, the question we ask in the paper, \emph{did utilize all the imaging prior provided by MVPS well or can we do better?}. In this work, we show that by cleverly using multi-view image prior, we can perform better than UA-MVPS \cite{kaya2022uncertainty}. To accomplish that, we introduce neural volume rendering method to MVPS.





\subsection{Neural Volume Rendering} \label{sec:volume_rendering_section}
Recent work on volume rendering techniques has shown outstanding results in learning scene representations from multi-view images \cite{mildenhall2020nerf}. Although such techniques are impressive with novel view synthesis, they can not faithfully provide the object's geometry from the learned volume density, leading to inaccurate and noisy reconstructions. Therefore, for our work, we use SDF-based volume rendering approach \cite{yariv2021volume} which models volume density as a function of the signed distance value as follows:

\begin{equation}\label{eq:sdf_volume_density}
\begin{aligned}
  \sigma(\mathbf{x}) & = \alpha\mathbf{\Phi}_{\beta}(-f_{\theta}(\mathbf{x})), \\ ~\text{where} ~ \mathbf{\Phi}_{\beta}(s) & = 
  \begin{cases}
    \frac{1}{2} \text{exp}\big(\frac{s}{\beta}\big), & \text{if} ~s \leq 0 \\
    1 - \frac{1}{2} \text{exp} \big(-\frac{s}{\beta}\big),   & \text{if} ~ s > 0 
    \end{cases}
    \end{aligned}
\end{equation}
Here, $\alpha, \beta > 0$ are trainable parameters and $\mathbf{\Phi}_{\beta}(.)$ is the cumulative distribution function of a zero-mean Laplace distribution. Eq:\eqref{eq:sdf_volume_density} ensures a smooth transition of density values near the object boundary, and at the same time allows a suitable extraction of zero level set after optimization for surface recovery.    
Inspired by the classical volume rendering techniques \cite{kajiya1984ray, max1995optical}, the expected color $I(\mathbf{c}_i, \mathbf{v}_i)$ of a camera ray $\mathbf{x}_i(t) = \mathbf{c}_i + t\textbf{v}_i$ with camera center $\mathbf{c}_i \in \mathbb{R}^3$ and viewing direction vector $\textbf{v}_i \in \mathbb{R}^3 $ can be modeled as:
\begin{equation}\label{eq:volume_rendering_integral}
I(\mathbf{c}_i, \mathbf{v}_i) = \int_{t_n}^{t_f} T\big(\mathbf{x}_i(t)\big) \sigma\big(\mathbf{x}_i(t)\big)  \mathbf{r}_{\psi}\big( \mathbf{x}_i(t), \mathbf{n}_i(t), \mathbf{v}_i\big)  dt,
\end{equation}
where $T\big(\mathbf{x}_i(t)\big) = \text{exp} \big(-\int_{0}^{t}  \sigma(\mathbf{x}_i(s)) ds \big)$ is the transparency, $\mathbf{n}_i(t) = \nabla_{\mathbf{x}} f_{\theta}(\mathbf{x}_i(t))$ is the level set's normal at $\mathbf{x}_i(t)$, $\mathbf{r}_{\psi}$ is the radiance field function and ($t_n$, $t_f$) are the bounds of the ray. Using the quadrature rule for numerical integration \cite{max1995optical} and the ray sampling strategy in \cite{yariv2020multiview}, we approximate the expected color as : 
\begin{equation}\label{eq:volume_rendering_integral}
 \hat{I}(\mathbf{c}_i, \mathbf{v}_i) = \sum_{j\in \mathcal{S}_i}^{} T_j \big(1 - \text{exp}(-\sigma_j\delta_j)\big) \mathbf{r}_{\psi}\big( \mathbf{x}_j, \mathbf{n}_j, \mathbf{v}\big)
\end{equation}
Here, $\mathcal{S}_i$ is the set of samples along the ray, $\delta_j$ is the distance between each adjacent samples and $T_j$ is the approximated transparency \cite{yariv2020multiview}. To realize $\mathbf{r}_{\psi}$, we introduce a second MLP with learnable parameters $\psi$. The radiance fields network $\mathbf{r}_{\psi}$ is placed subsequent to the signed distance field network $f_{\theta}$ (see Fig. \ref{fig:pipeline}). 
Furthermore, we introduce a feature vector $\mathbf{z} \in \mathbb{R}^{256}$ that is extracted from $f_{\theta}$ using a fully connected layer. This feature vector is fed to the radiance field network $r_{\psi}$ to account for global illumination effects. We optimize $f_{\theta}$ and $\mathbf{r}_{\psi}$ network on the test subject together. After optimization, we extract the zero level set of $f_{\theta}$ and recover the shape mesh using marching cubes algorithm \cite{lorensen1987marching}.
For more details, refer to Sec.\S \ref{sec:implementation_details} and \cite{yariv2021volume}. 


\smallskip
\formattedparagraph{Optimization.}
Our overall training loss is as follows:
\begin{equation}\label{eq:overall_loss}
    \begin{aligned}
  \mathcal{L}_{\text{mvps}} & =  \frac{1}{\mathcal{I}}\sum_{i=1}^{\mathcal{I}}  \Big( \overbrace{\textcolor{blue!80!black}{{c_{i}^{\text{mvs}}}}|f_{\theta}(\mathbf{p}_{i})|}^{\textrm{MVS Loss}} + \overbrace{\textcolor{green!70!black}{{c_{i}^{\text{ps}}}} \lVert \mathbf{n}_{i}^{r} - \mathbf{n}_{i}^{ps} \rVert}^{\textrm{PS Loss}} \\ & + \overbrace{(1-\textcolor{blue!80!black}{c_{i}^{\text{mvs}}}\textcolor{green!70!black}{c_{i}^{\text{ps}}})\lVert I_i - \hat{I}(\mathbf{c}_i, \mathbf{v}_i)\rVert_{1}}^{\textrm{Rendering Loss}} \Big) & \\
  + \frac{\lambda_m}{|\mathcal{M}|} & \sum_{i\in \mathcal{M}}^{} \overbrace{ CE\big(\max_{j \in \mathcal{S}_i} (\sigma_j / \alpha ), 0 \big)}^{\text{Mask Loss}} 
   + \overbrace{\lambda_e \mathbb{E}_{\mathbf{x}}  ( \lVert \nabla_{\mathbf{x}}f_{\theta}(\mathbf{x})\rVert -1 )^2}^{\textrm{Eikonal Regularization}} 
      \end{aligned}
\end{equation}
Eq.\eqref{eq:overall_loss} consists of five terms. Here, the first term forces the signed distance to vanish on the high fidelity position predictions of deep-MVS network. Similarly, the second term encourages the expected surface normal on a ray $\mathbf{n}_i^r = \sum_{j\in \mathcal{S}_i}^{} T_j (1 - \text{exp}(-\sigma_j\delta_j)) \textbf{n}_i(t)$ to align with the highly confident deep-PS predictions. The third term introduces an uncertainty-aware rendering loss to the optimization for the pixels where either MVS or PS fails. Intuitively, this allows the optimization to recover the missing surface details using rendering. 
We further improve the geometry using the object masks. For that, we first find the maximum density on rays outside the object mask (i.e. $i \in \mathcal{M}$). Then, we apply cross-entropy loss (CE) to minimize ray and geometry intersections as in \cite{yariv2020multiview}. The final term applies Eikonal regularization for plausible surface recovery as discussed in Sec.\S \ref{sec:IGR}. Fig.(\ref{fig:pipeline}) shows the overall pipeline of our proposed approach.





\section{Experiment and Results}

\smallskip
\formattedparagraph{Datasets.}
First, we evaluated our approach on the DiLiGenT-MV \cite{li2020multi}. DiLiGenT-MV is a standard benchmark for the MVPS setup, consisting of five real-world objects. The images are acquired using a turntable setup where the object is placed $\sim 1.5 m$ away from the camera. The turntable is rotated with 20 uniform rotations for each object, and 96 distinct light sources are used to capture light-varying images at each rotation. Although the DiLiGenT-MV benchmark consists of challenging objects with non-Lambertian surfaces, all provided objects satisfy isotropic BRDF property. Therefore, we simulated a new dataset consisting for objects with anisotropic and glossy surfaces. 

Similar to classical setup, we simulated our dataset using a turntable setup with 36 angle rotations. We place 72 light sources in a concentric way around the camera (see Fig.\ref{fig:acquisition_setup}) and rendered images corresponding to each light source.
We use licensed Houdini software to simulate our setup and render MVPS images of a single object 3D model taken from NeRF synthetic dataset \cite{mildenhall2020nerf} with three different material types (Wood, Gray, Red)\footnote{CC-BY-3.0 license.}. 
The Wood category is rendered to study anisotropic material behavior and the other two categories to analyse our method's performance on texture-less glossy objects. We rendered images at $1280 \times 720$ resolution to better capture the object details\footnote{Our dataset and further details related to it will be available soon.}.

\subsection{Implementation Details} \label{sec:implementation_details}
We implemented our method in Python 3.8 using PyTorch 1.7.1 \cite{paszke2017automatic} and conducted all our experiments on a single NVIDIA GPU with 11GB of RAM.  
We first train uncertainty-aware deep-MVS and deep-PS networks under a supervised setting. Then, we use these networks to have 3D position and surface normal predictions at test time. Finally, MVS images, along with the network predictions and their per-pixel confidence values, are used to optimize the proposed loss function (Eq.\eqref{eq:overall_loss}).



\smallskip
\formattedparagraph{(a) Deep-MVS Network.}
The deep-MVS network is trained on DTU's train set \cite{aanaes2016large}. The training takes 8 epochs using the learning rate $0.001$ and Adam optimizer \cite{kingma2014adam}. We use the MVS trained model at three coarser stages at test time to predict depth with coarse-to-fine approach. The depth $\mathbf{d}_i$ and the confidence $\mathcal{C}_i$ at each pixel $i$ are computed using Eq:\eqref{eq:depth regression}. The predicted depth is further enhanced using \cite{hui2016depth} work and converted to a set of 3D points $\{\mathbf{p}_{i}\}_{i=1}^{\mathcal{I}}$ by back-projecting the depth values to 3D space. Finally, we obtain binary confidences $\textcolor{blue!80!black}{{c}_i^{\text{mvs}}}$ by setting $\tau_{\text{mvs}} = 0.9$ for reliable fusion of confident position predictions.

\begin{table*}[t]
\small
\centering
    \resizebox{1.0\textwidth}{!}
    {
    \begin{tabular}{c|c|c|c|c|c|c|c|>{\columncolor{orange!20}}c}

   \textbf{ Method Category $\rightarrow$} &
   \multicolumn{2}{c|}{\cellcolor{blue!20} Deep Multi-View Stereo} &
     \multicolumn{3}{c|}{\cellcolor{green!20} Photometric Stereo}  &
   \multicolumn{2}{c|}{\cellcolor{red!20} View-Synthesis} 
    \\ 
    \hline
       \textbf{Dataset}$\downarrow$ $|$ \textbf{Method} $\rightarrow$ &
       \cellcolor{blue!20} MVSNet \cite{yao2018mvsnet} &
       \cellcolor{blue!20} PM-Net \cite{wang2021patchmatchnet} &
          \cellcolor{green!20} Robust PS \cite{oh2013partial} &
       \cellcolor{green!20} SDPS-Net \cite{chen2019self} &
       \cellcolor{green!20} CNN-PS \cite{ikehata2018cnn} &
       \cellcolor{red!20} NeRF \cite{mildenhall2020nerf} &
       \cellcolor{red!20} VolSDF \cite{yariv2021volume} &
       \textbf{~~Ours~~}  \\
        \hline
        BEAR     & 0.135 & 0.672 & 0.266 & 0.239 & 0.293 & 0.865 & 0.962  & \textbf{0.965}   \\
        BUDDHA   & 0.147 & 0.799 & 0.367 & 0.298 & 0.363 & 0.713 & 0.786  & \textbf{0.993}   \\
        COW      & 0.095 & 0.734 & 0.245 & 0.447 & 0.511 & 0.810 & 0.985  & \textbf{0.987}  \\
        POT2     & 0.126 & 0.666 & 0.231 & 0.464 & 0.632 & 0.859 & 0.946  & \textbf{0.991}  \\
        READING  & 0.115 & 0.834 & 0.242 & 0.188 & 0.508 & 0.673 & 0.683  & \textbf{0.975}  \\
        \hline
        \textbf{AVERAGE} & 0.124  & 0.741 & 0.270 & 0.327 & 0.461 &  0.784 & 0.873 & \textbf{0.982}\\
    \hline
    \end{tabular}
    }
    \caption{ \textit{F}-score comparison of standalone method reconstructions on DiLiGenT-MV benchmark \cite{li2020multi}. Our method outperforms standalone multi-view stereo, photometric stereo and view synthesis methods in all of the object categories. 
    }
    \label{tab:standalone_numerical_comparison}
\end{table*}

\begin{figure}
    \centering
    \includegraphics[width=1.0\columnwidth]{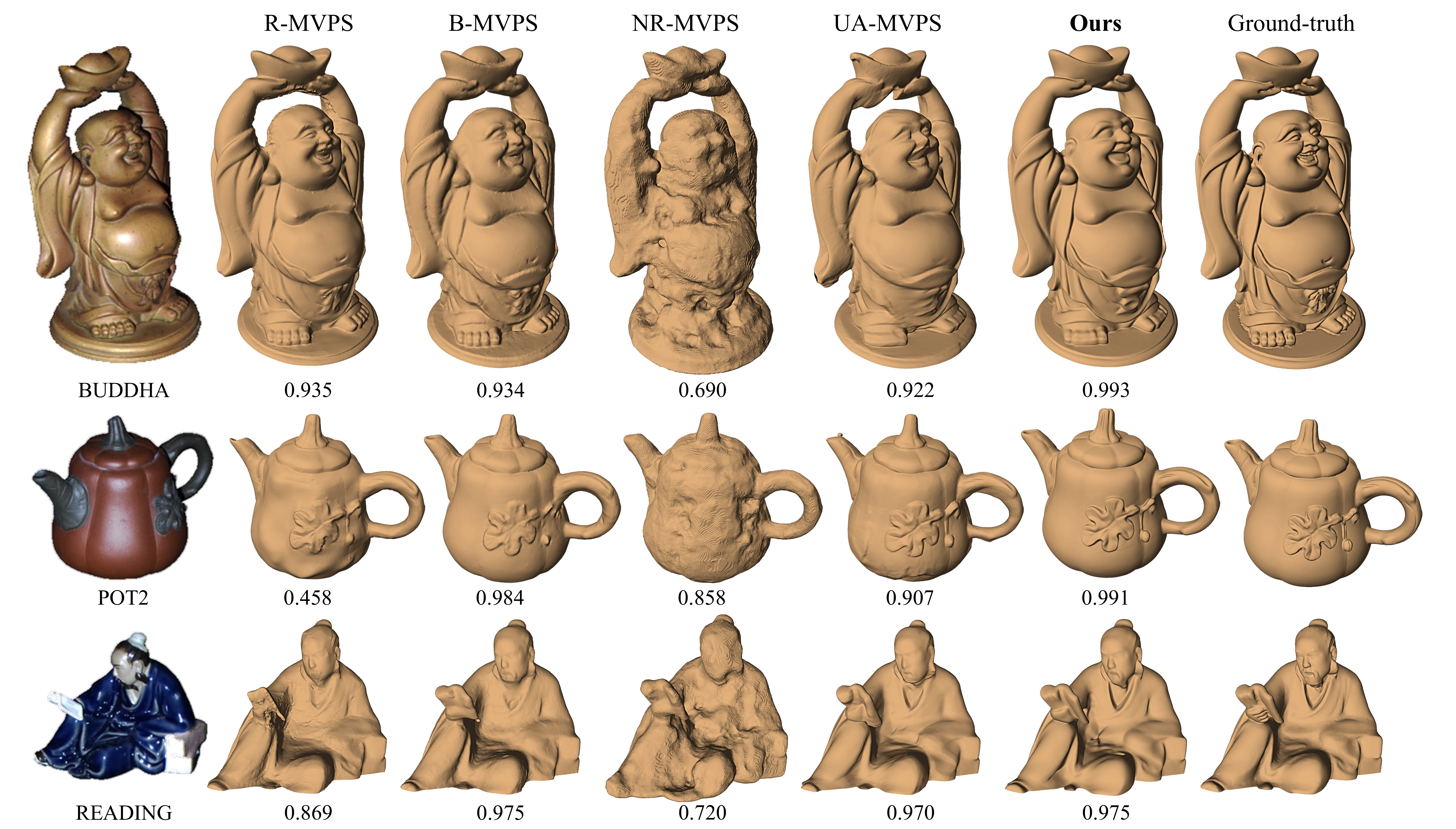}
    \caption{Comparison of MVPS reconstructions on DiLiGenT-MV benchmark \cite{li2020multi}. We report \textit{F}-score metric results for numerical comparison. We can observe that our method recovers fine details and provides high-quality reconstructions of challenging objects.   
    }\label{fig:diligent_results}
\end{figure}

\smallskip
\formattedparagraph{(b) Deep-PS Network.} 
We train the deep-PS network on CyclesPS dataset \cite{ikehata2018cnn} for 10 epochs using Adam optimizer \cite{kingma2014adam} and learning rate of $0.1$. We use probability of $p_{\text{mc}} = 0.2$ in every dropout layer of the architecture. For training, we set $\mathbf{N}_{\text{mc}} = 10$ and $\lambda_w = 10^{-4}$  (see Eq:\eqref{eq:ps_network_loss}). At test time, we first create observation map per-pixel using MVPS images. We then run the network on each observation map 100 times to have the output surface normal $\mathbf{n}_i^{\text{ps}}$ and the prediction variance $\tilde{\sigma}_i^{2}$ \cite{gal2015bayesian, gal2016dropout}. Finally, we obtain the confidence value $\textcolor{green!70!black}{c_i^{\text{ps}}}$ at $i^{th}$ pixel by setting $\tau_{\text{ps}} = 0.03$.

\smallskip
\formattedparagraph{(c) Overall Shape Optimization.}
As described in \S Sec.\ref{sec:volume_rendering_section}, we optimize two networks during optimization: signed distance field network ($f_{\theta}$) and radiance field network ($\mathbf{r}_{\psi}$). 
$f_{\theta}$ consists of 8 MLP layers with a skip connection connecting the first layer to the $4^{th}$. On the other hand, $\mathbf{r}_{\psi}$ has four MLP layers (see Fig.\ref{fig:pipeline}). All the layers of both networks have 256 units. 
We apply Fourier feature encoding to the inputs (position $\mathbf{x}$ and view direction $\mathbf{v}$) to improve the networks' ability to represent high-frequency details \cite{mildenhall2020nerf}.  
For the loss function in Eq:\eqref{eq:overall_loss}, we set $\lambda_m = 0.1$ and $\lambda_e = 1$.
We use a set of multi-view images which are captured under the illumination of the same randomly chosen light source to compute the rendering loss.
We use Adam optimizer \cite{kingma2014adam} with learning rate $10^{-4}$ and train for $10^4$ epochs. In each epoch, we use batches of $1024$ rays from each view and sample 64 points along each ray \cite{yariv2021volume}.
To compute the Eikonal regularization as in Eq:\eqref{eq:loss_eikonal}, we also uniformly sample points globally. So, the distribution $\mathcal{D}$ stands for the collection of these ray samples and global samples. 
After the optimization, we extract zero level set of the learned SDF representation by $f_{\theta}$ and recover the shape mesh using marching cubes algorithm \cite{lorensen1987marching} on a $512^3$ grid.

 \begin{figure*}[t]
\centering
\subfigure[\label{fig:synthetic_data_bar} Chamfer $L_2$ comparison]{\includegraphics[width=0.25\linewidth]{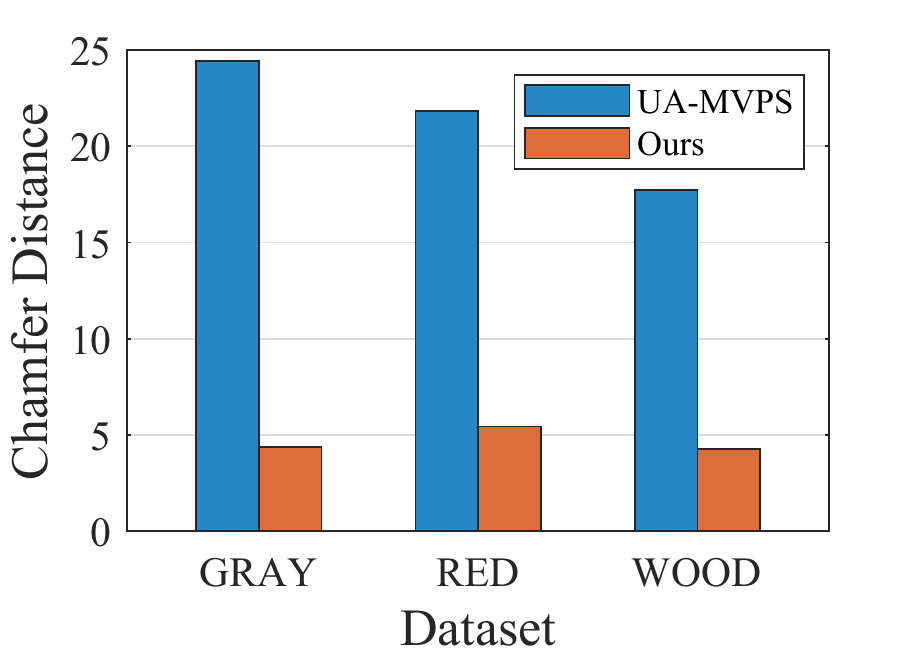}} 
\subfigure[\label{fig:prediction_analysis} Texture-less object ]{\includegraphics[width=0.37\linewidth]{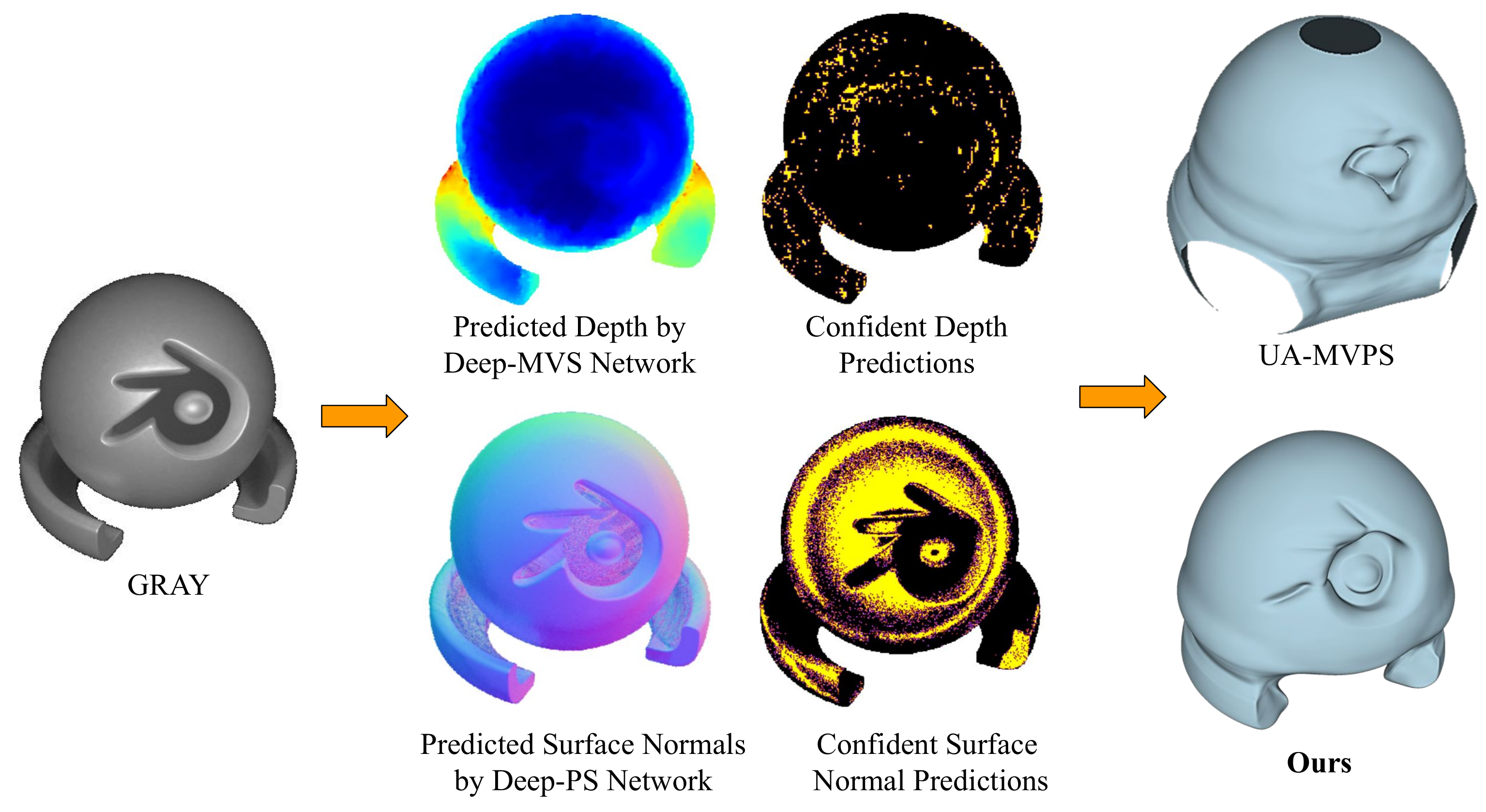}}
~~~~~\subfigure[\label{fig:tsdf_fusion} Comparison with TSDF-Fusion]{\includegraphics[width=0.33\linewidth]{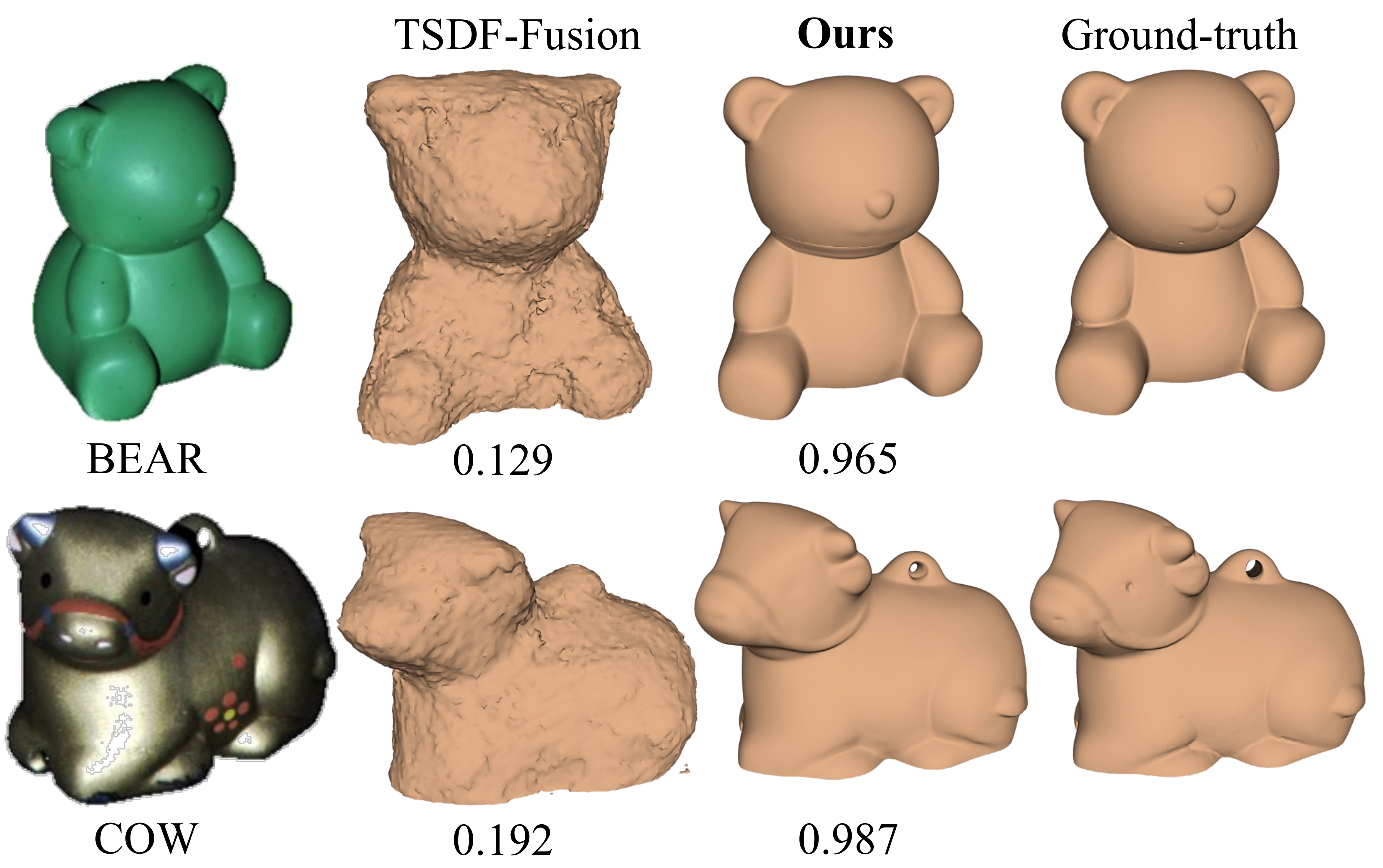}}
\caption{(a) Chamfer $L_2$ comparison of our method with UA-MVPS \cite{kaya2022uncertainty} on our synthetic dataset (lower is better).
(b) We show depth and surface normal predictions on texture-less object. Pixels marked with yellow color indicate confident MVS or PS predictions ($\textcolor{blue!80!black}{c_{i}^{\text{mvs}}}$ and ${\textcolor{green!70!black}{c_{i}^{\text{ps}}}}$). Note that MVS cannot predict depth reliably on texture-less surface, which leads to inferior results in UA-MVPS \cite{kaya2022uncertainty}. On the other hand, our uncertainty-aware volume rendering approach can recover missing surface information, and therefore, provides better reconstructions. (c) Comparison of our method with TSDF Fusion algorithm \cite{curless1996volumetric}. We report \textit{F}-score metric for numerical comparison. 
}
\label{fig:synthetic_data_results}
\end{figure*}

\subsection{Statistical Analysis}
We performed comparative analysis on the DiLiGenT-MV dataset \cite{li2020multi}. 
To evaluate the quality of the shape reconstructions, we use well-known Chamfer-$L_2$ and \textit{F}-score\cite{knapitsch2017tanks} metric. For better understanding, we present the performance comparison result in two different categories depending on the method type.

\smallskip
\formattedparagraph{(a) Standalone Method Comparison.} By the standalone method, we refer to the approaches that use only one modality \ie, either MVS or PS images for 3D reconstruction. We consider SOTA MVS, PS, and view-synthesis methods for this comparison. Note that we use Horn and Brooks algorithm \cite{horn1986variational} for normal integration to recover depth maps. We then back-project the recovered depths to 3D space to evaluate reconstruction performance. Table \ref{tab:standalone_numerical_comparison} presents the \textit{F}-score comparison of these methods on DiLiGenT-MV \cite{li2020multi}. 
The statistics show that our method consistently outperforms the standalone approaches. Further, we observed that none of the standalone methods could reliably recover the object's 3D shape. On the contrary, our method gives accurate reconstruction by effectively exploiting the complementary surface and image priors.

\smallskip
\formattedparagraph{(b) MVPS Methods Comparison.}
Table \ref{tab:mvps_numerical_comparison} provides the \textit{F}-score comparison results with SOTA MVPS methods on the DiLiGenT-MV benchmark dataset. For our comparison, we consider both explicit geometry modeling-based classical approaches \cite{park2016robust,li2020multi}, and neural implicit representation based deep approaches \cite{kaya2021neural, kaya2022uncertainty}. The numerical results show that our method provides the highest scores on three objects categories. Moreover, it outperforms all the existing MVPS methods on average. Some important point to note is that \textit{\textbf{(i)}} Our approach provides a scalable and easy-to-execute implementation, without requiring tedious sequential steps as in classical methods \cite{li2020multi}, \textit{\textbf{(ii)}} Our MLP based shape representation requires only 3.07MB of memory, while explicit geometric methods may require up to 90MB. Such advantages make our method an efficient and effective algorithmic choice for solving MVPS.

%
%
%
%
%
%

\begin{table}[t]
    \centering
    \resizebox{\columnwidth}{!}
    {
    \begin{tabular}{c|c|c|c|c|>{\columncolor{orange!20}}c}
    \hline
       \textbf{Dataset}$\downarrow$ $|$ \textbf{Method} $\rightarrow$  & R-MVPS \cite{park2016robust} & B-MVPS \cite{li2020multi} & NR-MVPS \cite{kaya2021neural} &  UA-MVPS \cite{kaya2022uncertainty} & \textbf{Ours}\\
        \hline
        BEAR     & 0.504 & \textbf{0.986} & 0.856 & 0.895  & 0.965   \\
        BUDDHA   & 0.935 & 0.934 & 0.690 & 0.922 & \textbf{0.993}  \\
        COW      &  0.915 & \textbf{0.989} &0.844 & 0.979 & 0.987   \\
        POT2     &  0.458 & 0.984 & 0.858 & 0.907 & \textbf{0.991}   \\
        READING  &  0.869 & 0.975 &0.720 & 0.970 & \textbf{0.975}   \\
        \hline
        \textbf{AVERAGE} & 0.736 & 0.974 & 0.794 & 0.935 & \textbf{0.982}    \\
    \hline
    \end{tabular}
    }
    \caption{\textit{F}-score comparison of MVPS reconstructions on DiLiGenT-MV benchmark \cite{li2020multi}. Our method performs consistently well on various objects and is better than others on average.}
    \label{tab:mvps_numerical_comparison}
\end{table}

\begin{figure}[t] 
\centering
{\includegraphics[width=1\linewidth]{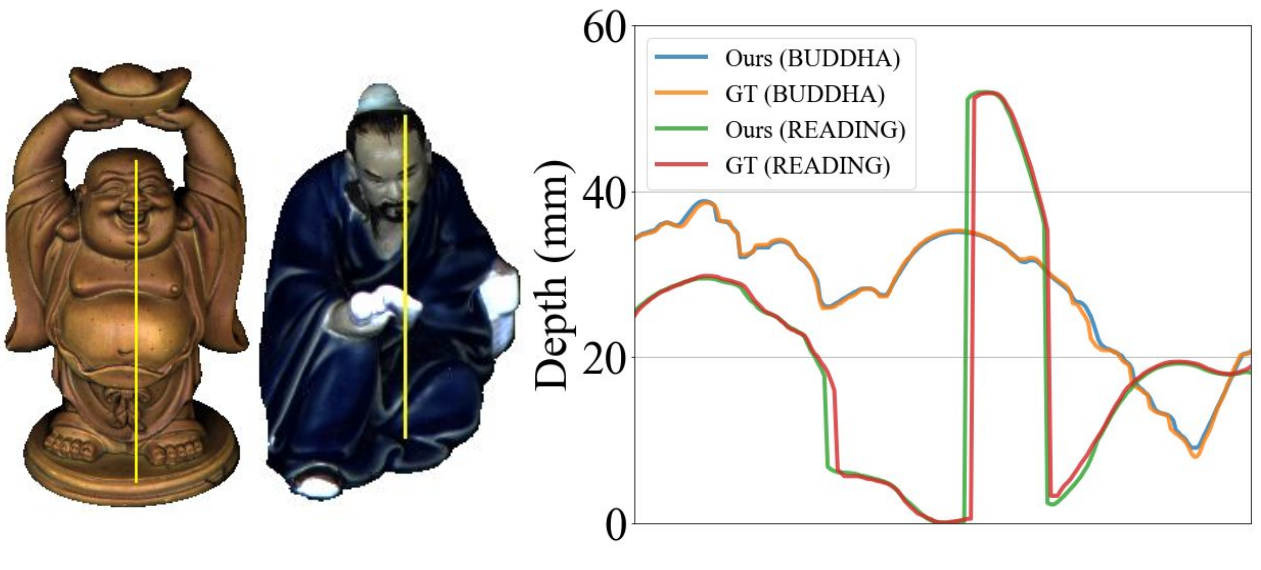}}
\caption{Surface profile of our reconstructions on a randomly chosen path. Clearly, our surface profile overlaps with the ground-truth(GT), which indicates the high quality of our reconstructions. 
}
\label{fig:surface_profile}
\end{figure}

\subsection{Further Analysis}
\smallskip
\formattedparagraph{(a) Anisotropic and Textureless Glossy Surfaces.} 
We perform evaluations on our synthetic dataset to analyze the efficiency of our approach on anisotropic and texture-less glossy surfaces. In Fig.\ref{fig:synthetic_data_bar}, we provide Chamfer $L_2$ metric comparison of our method with the recent UA-MVPS \cite{kaya2022uncertainty}. The results show that our method performs much better than its competitor on glossy (Gray, Red) and anisotropic surfaces (Wood). In Fig.\ref{fig:prediction_analysis}, we show qualitative results of the uncertainty-aware deep-MVS and deep-PS networks on the Gray category. It can be observed from visual results that deep-MVS cannot provide reliable position estimates on texture-less glossy surfaces. For this reason, methods relying on the fusion of only MVS and PS priors (such as UA-MVPS) cannot handle all kinds of surfaces. On the other hand, our method can recover the missing surface information by effectively utilizing volume rendering; hence, it can suitably work for anisotropic and glossy surface profiles. 

\smallskip
\formattedparagraph{(b) Optimization.}
Here, we investigate the effectiveness of our proposed optimization loss in Eq:\eqref{eq:overall_loss} with an ablation study. For that, we compare the reconstruction quality of our method by removing (\textit{i}) MVS loss term, (\textit{ii}) PS loss term, (\textit{iii}) rendering loss term and (\textit{iv}) uncertainty modeling ($\textcolor{blue!80!black}{c_{i}^{\text{mvs}}}$ and ${\textcolor{green!70!black}{c_{i}^{\text{ps}}}}$) from the overall loss. In Table \ref{tab:ablation}, we provide Chamfer $L_2$ metric comparison of the reconstruction quality achieved under each of these configurations. The numerical results verify that uncertainty modeling based integration of MVS, PS and rendering loss terms provides best results on DiLiGenT-MV \cite{li2020multi}.

\smallskip
\formattedparagraph{(c) Surface Profile.}
To show the quality of our recovered 3D reconstructions, we study the surface topology across an arbitrarily chosen curve on the surface. Fig.\ref{fig:surface_profile}
shows a couple of examples of such surface profile on Buddha and Cow sequences. Clearly, our recovered surface profiles align well with the ground truth.

\smallskip
\formattedparagraph{(d) Volumetric Fusion Approach.}  
Of course, one can use robust 3D fusion method such as TSDF fusion \cite{curless1996volumetric} to recover the object's 3D reconstruction. And therefore, we conducted this experiment to study the results that can be recovered using such fusion techniques. Accordingly, we fuse
deep-MVS depth and the depth from deep-PS normal integration \cite{horn1986variational} using the TSDF fusion. Fig.\ref{fig:tsdf_fusion} shows that TSDF fusion provide inferior results compared to ours.

\begin{table}
    \centering
    \resizebox{\columnwidth}{!}{
		\begin{tabular}{c|c|c|c|c|c|c}
			\hline
			\textbf{Settings}$\downarrow$ $|$ \textbf{Dataset} $\rightarrow$ & ~BEAR~ & ~BUDDHA~ & ~COW~  & ~POT2~ & ~READING~ & ~\textbf{AVERAGE}~ \\ \hline
			w/o MVS Loss & 0.189 &	0.089 &	0.202 &	0.156 &	0.353 &	0.198  \\ \hline
			w/o PS Loss &  0.301 &	0.572& 0.184 &	0.262 &	0.428 &	0.349  \\ \hline
			w/o Rendering Loss & 0.154	& 0.471 &	0.269 &	0.235 &	0.374 &	0.301 \\ 	\hline
			w/o Uncertainty-Aware. &  0.267 &	0.085 &	0.313 &	0.137 &	0.251 &	0.211 \\ 	\hline
		\rowcolor{orange!20}	Ours  & 0.213 &	0.088 &	0.176 &	0.198 &	0.253 &	\textbf{0.186}	  \\ \hline
		\end{tabular}}
		\caption{ Contribution of MVS, PS, rendering loss terms and uncertainty modeling to our reconstruction quality. We report Chamfer $L_2$ metric for comparison (lower is better). Clearly, our proposed loss in Eq:\eqref{eq:overall_loss} produces best results on average.
		}
		\label{tab:ablation}
\end{table}

\smallskip
\formattedparagraph{(e) Limitations.}
Although our method works well on glossy objects, it may fail on materials with mirror reflection. Furthermore, SDF representation of the object shape restricts our approach to solid and opaque materials. Finally, our work considers a calibrated setting for MVPS setup, and it would be interesting to further investigate our approach in an uncalibrated setup. \textit{For more results and exhaustive analysis of our method refer to our supplementary.}

\section{Conclusion}

The proposed method addresses the current limitations of well-known MVPS methods and makes it work well for diverse object material types. Experimental studies on anisotropic and texture-less glossy objects show that existing MVS and PS modeling techniques may not always extract essential cues for accurate 3D reconstructions. However, by integrating incomplete yet reliable MVS and PS information into a rendering pipeline and leveraging the generalization ability of the modern view synthesis approach to model complex BRDFs, it is possible to make MVPS setup work well for anisotropic materials and glossy texture-less objects with better accuracy. Finally, the performance on the standard benchmark shows that our method outperforms existing methods providing exemplary 3D reconstruction results. To conclude, we believe that our approach will open up new avenues for applying MVPS to real-world applications such as metrology, forensics, etc.

\formattedparagraph{Acknowledgement.} {The authors thank ETH support to the Computer Vision Lab (CVL) and Focused Research Award from Google (ETH 2019-HE-318, 2019-HE-323, 2020-FS-351, 2020-HS-411).}



{\small
\balance
\bibliographystyle{ieee_fullname}
\bibliography{egbib}
}

\end{document}